\documentclass[conference]{IEEEtran}
\IEEEoverridecommandlockouts
\usepackage{caption}
\usepackage{cite}
\usepackage{amsmath,amssymb,amsfonts}
\usepackage{algorithmic}
\usepackage{graphicx}
\usepackage{textcomp}
\usepackage{xcolor}
\usepackage{subcaption}
\usepackage{cite}
\usepackage{float}
\usepackage{enumitem}
\usepackage{multirow}
\def\BibTeX{{\rm B\kern-.05em{\sc i\kern-.025em b}\kern-.08em
    T\kern-.1667em\lower.7ex\hbox{E}\kern-.125emX}}

\begin{document}
\raggedbottom

\title{Explainability, risk modeling, and segmentation  based customer churn analytics for personalized retention in e-commerce}

\IEEEoverridecommandlockouts

\author{
  \IEEEauthorblockN{Indrajith Ekanayake and Sanjula De Alwis}
  \IEEEauthorblockA{School of Computing, Informatics Institute of Technology, Sri Lanka\\
  indrajith.e@iit.ac.lk, pramuditha.20232457@iit.ac.lk}\thanks{979-8-3315-5723-2/26/\$31.00 ©2026 IEEE}
}

% \author{
% \IEEEauthorblockN{Sanjula De Alwis}
% \IEEEauthorblockA{\textit{School of Computing} \\
% \textit{Informatics Institute of Technology}\\
% Sri Lanka \\
% pramuditha.20232457@iit.ac.lk}
% \and
% \IEEEauthorblockN{Indrajith Ekanayake}
% \IEEEauthorblockA{\textit{School of Computing} \\
% \textit{Informatics Institute of Technology}\\
% Sri Lanka \\
% indrajith.e@iit.ac.lk}
% }

\maketitle

\begin{abstract}
In online retail, the cost to attract new customers is higher than retaining existing customers. As a result, businesses invest in predictive analytics for customer churn. However, modern churn prediction algorithms are black boxes that lack the transparency required to identify what actually causes attrition, tenure-based retention, and customer segments with a high risk of churn. Therefore, the next generation of churn analytics focus must shift from prediction to personalized retention. This study proposes a threefold approach that combines explainable AI to describe feature importance insights, survival analysis to model time-to-event data, and finally RFM (recency, frequency, and monetary) approach to segment customers based on transactional behavior. With this explainability, risk modeling, and segmentation-based approach, personalized retention strategies were proposed to reduce attrition and increase customer loyalty. We propose that by following this approach, organizations can determine why churn is likely, when intervention is most effective, and which customer groups to prioritize. 
\end{abstract}

\vspace{0.5em}
\renewcommand\IEEEkeywordsname{Keywords}

\section{Introduction}
Electronic commerce integrates internet-based information technologies with conventional offline business operations. After experiencing several golden ages, today, e-commerce has matured into a highly competitive market. Surviving in this competitive environment requires taking full use of consumer data and formulating personalized business strategies. Reducing customer attrition has gained much recent attention, as the cost of acquiring new customers often exceeds the cost of retaining existing ones \cite{ind19}. More specifically, studies such as Çelik and Osmanoğlu report that acquiring new customers can cost up to ten times more than retaining existing customers\cite{ind20}. 

E-commerce customer attrition has been examined using a range of analytical approaches, including machine learning \cite{ind13,ind21,ind22}, and deep learning techniques \cite{ind24,ind25}. However, highly accurate machine-learning models often lack interpretability \cite{ind26}, functioning as binary classification black boxes that obscure the contribution of individual features. Also, customer behaviors are highly dynamic, influenced by factors such as seasonality, product lifecycle, and external market conditions, making static churn prediction models less effective. Therefore, techniques such as Kaplan-Meier analysis \cite{ind22}, and Cox proportional hazards regression \cite{ind23} are also increasingly being adopted for time-to-event modeling. E-commerce studies have also used churn prediction in conjunction with customer segmentation. Notably, segmentation based on value and behavior has been identified as the most widely adopted segmentation technique in e-commerce \cite{ind27}. 

Despite substantial work on customer churn analytics, the literature remains fragmented: studies often treat tasks such as churn prediction and segmentation in isolation rather than as an integrated pipeline \cite{ind28}. To the best of the authors’ knowledge, this is the first work to integrate predictive accuracy, model interpretability, temporal modeling, segmentation, and real-time deployment within a single unified system, thereby enabling personalized analytics for e-commerce. The main contributions of this work are as follows:
\begin{itemize}
    \item Identify the best performing classification model for the dataset using the high recall, stability between training and testing performance, and low standard deviation across cross-validation folds.
    \item Combining the best performing predictive model with explainability, risk modeling, and segmentation analytics for the proposal of personalized retention strategies.
\end{itemize}

This paper is organized as follows. Section II reviews related work. Section III describes the methodology and experimental setup. Section IV analyzes the results, including explainability, risk modeling, and segmentation-based customer churn analysis. Section V summarizes the contributions and suggests directions for future research.

\section{Literature Review}
Customer churn prediction is a widely studied phenomenon across industries such as telecommunications \cite{ind16, ind30}, e-commerce \cite{ind28, ind25, ind24, ind21, ind22, ind12, ind13}, banking \cite{ind32}, online gaming \cite{ind15, ind31}, and logistics \cite{ind29}. Early churn prediction approaches relied primarily on rule-based systems and conventional statistical models. Rule-based systems use expert-defined heuristics based on simple business metrics like recency, frequency, and monetary value (RFM analysis) \cite{ind17, ind18}. These systems were transparent and easily interpretable, allowing marketers to define and adjust rules manually. However, in e-commerce, customer attrition is influenced by a number of variables ranging from specific consumer characteristics to market dynamics. Therefore, the focus was shifted towards machine learning and deep learning approaches for higher accuracy. Matuszelański and Kopczewska \cite{ind21} used binary regression models: logistic regression and extreme gradient boosting (XGBoost), identifying churn/non-churn behaviour. These models can give better performance, but at the cost of losing direct explainability. This has been more evident in the attempts to use deep learning for customer churn prediction in e-commerce. Deep learning models such as recurrent neural networks (RNN)  \cite{ind25, ind34},  convolutional neural networks (CNN) \cite{ind35}, transformer architectures \cite{ind33}, and multimodel approaches \cite{ind35, ind25}, showed promise in capturing sequential and complex patterns but faced challenges in interpretability, computational cost, and mainstream adoption. 

To address the gap in model interpretability, researchers have been testing out different approaches in isolation. Among these, explainable AI (XAI) has emerged as a prominent approach balancing accuracy with interpretability \cite{ind36}. Among XAI approaches shapley additive explanations (SHAP) and local interpretable model-agnostic explanations (LIME) were commonly adopted. Asif et al. \cite{ind36} used SHAP and LIME to quantify feature contributions to churn risk. They found that "revenue" and "regularity" were positively associated with churn (+0.21 and +0.06, respectively), whereas "apru\_segment" was negatively associated (-0.03), reducing the predicted probability of churn. Similarly, Tao et al. \cite{ind37} propose the GXAI (XAI in Online Games) workflow and evaluated it on player-churn prediction. This workflow pairs model types with post-hoc explainers across data views: SHAP for tabular/portrait and behavior features, class-activation mapping (CAM/Grad-CAM) for client images, and graph explainers (GNNExplainer, GraphLIME) for social-graph models. In e-commerce applications, TreeSHAP is widely used because tree-based models, particularly XGBoost, often achieve strong predictive performance \cite{ind12,ind13}. 

Survival analysis is another approach for modeling time-to-event dynamics and incorporating censored data. This was originally developed in medical research to predict the lifetime expectancy of an individual, but more recently, this group of methods has also been applied in churn analytics research \cite{ind15, ind16}. Peri{\'a}{\~n}ez et al. \cite{ind15} used a survival ensemble model to predict both survival probability and time-to-churn of the players in online games. Masarifoglu and Buyuklu \cite{ind16} applied the survival techniques: Kaplan-Meier analysis and Cox proportional hazards model to quantify and describe the risk until customer churns in the telecom domain. Similarly, time-to-churn dynamics have been studied in financial services, telecommunications, online gaming, and digital libraries. However, to the best of the authors’ knowledge, no published survival analysis literature addresses e-commerce, leaving a significant gap in the literature. By integrating churn prediction, XAI, survival analysis, and transaction-based segmentation, this study aims to address the void in the literature and enable personalized retention strategies for e-commerce, with the goal of reducing customer attrition and improving loyalty.

\section{Methodology}
In the paper authors used an open-access dataset to study e-commerce customer behavior and churn \cite{ind1}. During this process, five ML models were benchmarked for identifying the best-performing model. The process of developing personalized retention strategies in e-commerce is threefold. At first, explainable AI is used to describe feature importance insights. Secondly, survival analysis is performed by modeling time-to-event data. Lastly, the recency, frequency, and monetary (RFM) approach is used for customer segmentation based on transactional behavior. 

\subsection{Dataset Selection}\label{AA}
As discussed earlier, e-commerce churn prediction is less explored compared to industries like telecommunications and gaming, resulting in fewer publicly available benchmarking datasets \cite{ind2}. The dataset used in this study was sourced from Ankit Verma \cite{ind1} and has been employed in multiple prior studies \cite{ind3,ind4} due to its features and usability. This dataset contains transactional, behavioral, and demographic information of customers from a leading online e-commerce company. The company name or any PII information was not disclosed. The dataset includes 5,630 customer records with 20 features.

\subsection{Model Training and Analysis}\label{AA}
Initial data preprocessing involved exploratory data analysis (EDA) to prepare the dataset for modeling. During this phase, duplicate records were removed, and missing values were imputed. There were missing values in 7 features. Namely, \textit{DaySinceLastOrder}, \textit{OrderAmountHikeFromLastYear}, \textit{Tenure}, \textit{OrderCount}, \textit{CouponUsed}, \textit{HourSpendOnApp}, and \textit{WarehouseToHome}. Based on the analysis of interquartile ranges (Q1–Q3) and the presence of outliers, different imputation strategies were applied:
\begin{itemize}
    \item For features exhibiting a wide range and correlations with other variables (i.e., \textit{DaySinceLastOrder}, \textit{OrderAmountHikeFromLastYear}, \textit{Tenure}, \textit{OrderCount}, \textit{CouponUsed}, and \textit{WarehouseToHome}), missing values were imputed using the iterative imputer with a regression model\cite{ind7}. This approach models each feature with missing values as a function of others, iteratively estimating missing entries to better capture multivariate relationships.
    
    \item For \textit{HourSpendOnApp}, which has a smaller range, missing values were imputed using the median.
\end{itemize}

Categorical values were standardized using One-Hot Encoding. This method was chosen due to the limited number of categories per feature ($\leq$ 5 categories). Outlier detection was performed using the Mahalanobis Distance method, which measures the distance of each data point from the multivariate distribution center, considering feature covariance, with a significance threshold of \( p \leq 0.001 \) \cite{ind5}. The Chi-Squared test applied to the Mahalanobis distances identified 266 rows as outliers with a p-value less than 0.001. Consequently, these 266 rows were removed to prevent extreme or irrelevant values from distorting the model.

After EDA, the finalized dataset contained 4,807 records. The dataset was split into training and test sets using stratified sampling in an 80:20 ratio, ensuring that the class distribution of churn and non-churn customers was preserved in both subsets. 

\section{Results}

Multiple classification models were evaluated to identify the most effective approach to the prediction of customer churn. Models compared included Logistic Regression (LR), Decision Tree (DT), Random Forest (RF), XGBoost (XGB), and CatBoost (CB). The evaluation focused on three key criteria: high recall, stability between training and testing performance, and low standard deviation across cross-validation folds.

\begin{table}[H]
\centering
\caption{Model Performance Comparison}
\begin{tabular}{p{1cm}ccccc}
\hline
Model & Accuracy & Precision & Recall & F1 Score & Error Rate \\
\hline
XGB  & 0.964 & 0.847 & 0.952 & 0.896 & 0.036\\
DT       & 0.931 & 0.819 & 0.928 & 0.892 & 0.045 \\
RF       & 0.903 & 0.801 & 0.915 & 0.853 & 0.067 \\
CB & 0.892 & 0.798 & 0.881 & 0.922 & 0.089 \\
LR       & 0.883 & 0.821 & 0.471 & 0.571 & 0.091 \\
\hline
\end{tabular}
\label{tab:model_outlier_comparison}
\end{table}

Table~\ref{tab:model_outlier_comparison} presents the performance comparison of the evaluated models. Among all models, XGBoost stands out as the clear choice for customer churn prediction. Other models, Decision Tree, Random Forest, and CatBoost performed reasonably but with lower recall values. Logistic Regression presented the weakest performance. Based on this analysis, XGBoost was selected for explainability, risk modeling, and segmentation based customer churn analysis for personalized retention in e-commerce.

\begin{figure}[t]
    \centering
    \begin{subfigure}[t]{0.4\textwidth}
        \centering
        \includegraphics[width=\textwidth]{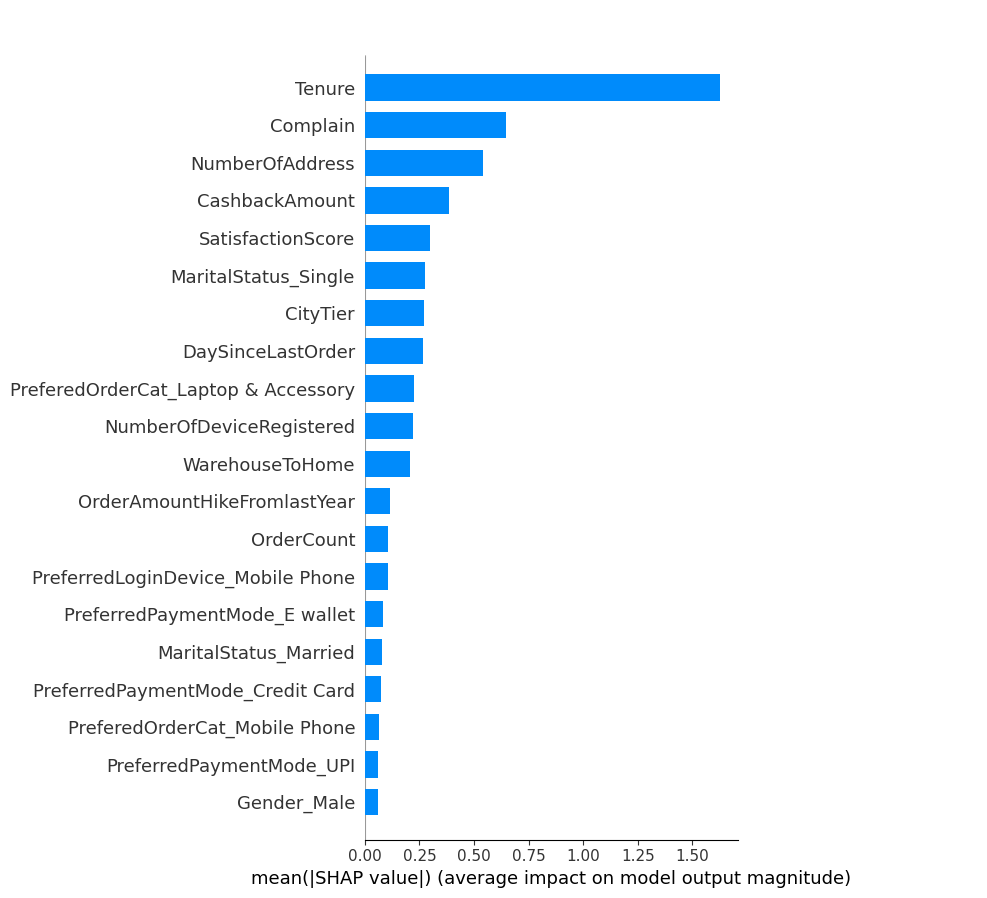}
        \caption{Feature importance on SHAP values}
        \label{fig:shap-values}
    \end{subfigure}
    \hfill
    \begin{subfigure}[t]{0.4\textwidth}
        \centering
        \includegraphics[width=\textwidth]{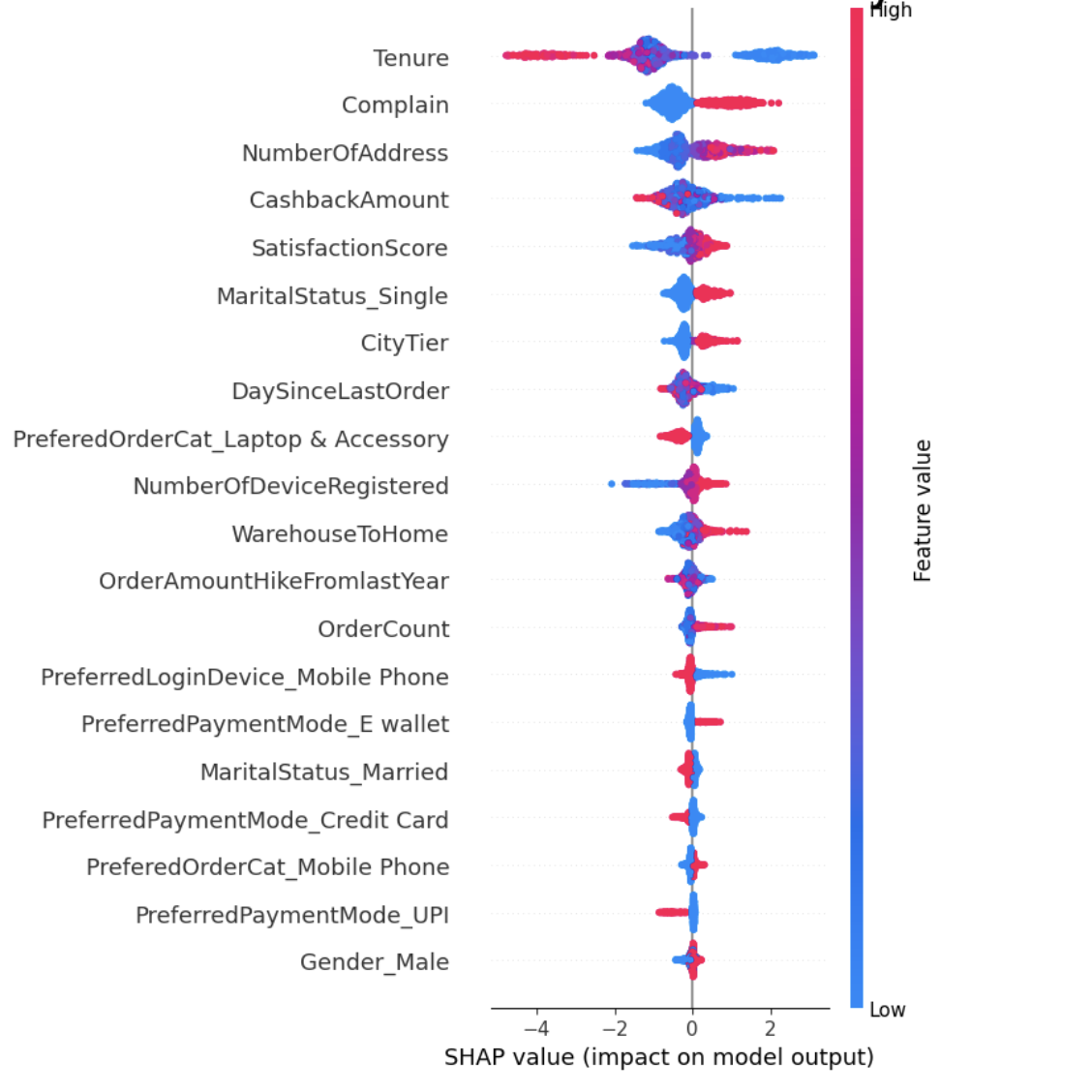}
        \caption{SHAP value distribution by feature}
        \label{fig:shap-feature}
    \end{subfigure}
    \caption{SHAP analysis}
    \label{fig:viewpointstemporal}
\end{figure}

\subsection{Explainable AI to Describe Feature Importance Insights}\label{AA}
SHapley Additive exPlanations (SHAP) has been one of the go-to interpretable machine learning frameworks. For tree-based models such as XGBoost, SHAP values admit efficient, model-exact computation via Tree SHAP, which is the most appropriate method for this class of models \cite{ind8,ind9,ind10,ind11}. Using Tree SHAP, we have computed feature attributions both globally (overall feature importance) and locally (individual prediction explanation). Post-hoc explanations of the churn model are presented at (Figs.~\ref{fig:shap-values} and~\ref{fig:shap-feature}). 

The global importance plot (Fig.~\ref{fig:shap-values}) summarizes, for each feature $j$, its average influence on the model output across all instances.

\begin{equation}
\mathrm{Imp}(j) = \frac{1}{n}\sum_{i=1}^{n}\lvert \phi_{ij} \rvert
\end{equation}

Here, $\phi_{ij}$ is the SHAP contribution of feature $j$ to instance $i$, and $n$ is the number of instances. A larger $\mathrm{Imp}(j)$ means the feature moves the prediction more on average. The SHAP beeswarm (Fig.~\ref{fig:shap-feature}) shows, for each feature, the distribution, direction, and magnitude of SHAP values. Based on direction and distribution, variables such as \textit{Tenure}, \textit{Complain}, \textit{NumberOfAddress}, and \textit{CashbackAmount} show clear dominance, whereas all other variables contribute materially less. Higher \textit{Tenure} is associated with negative SHAP values, indicating a lower propensity to churn. Customers who have lodged a \textit{Complain} show positive SHAP contributions, increasing churn risk. Larger \textit{CashbackAmount} and higher \textit{SatisfactionScore} tend to shift SHAP values below zero, suggesting a protective effect against churn. Longer gaps since the last order \textit{DaySinceLastOrder} and greater distance from \textit{WarehouseToHome} exhibit positive SHAP tails for many instances, implying elevated risk for those segments. Being single \textit{MaritalStatus\_Single} is weakly but directionally positive for churn in several cases, whereas \textit{NumberOfDeviceRegistered} shows a small negative association, consistent with deeper ecosystem engagement. Finally, payment-mode and category preferences, and gender, display tight SHAP spreads around zero.

\begin{figure*}[htbp]
\centerline{\includegraphics[width=\textwidth]{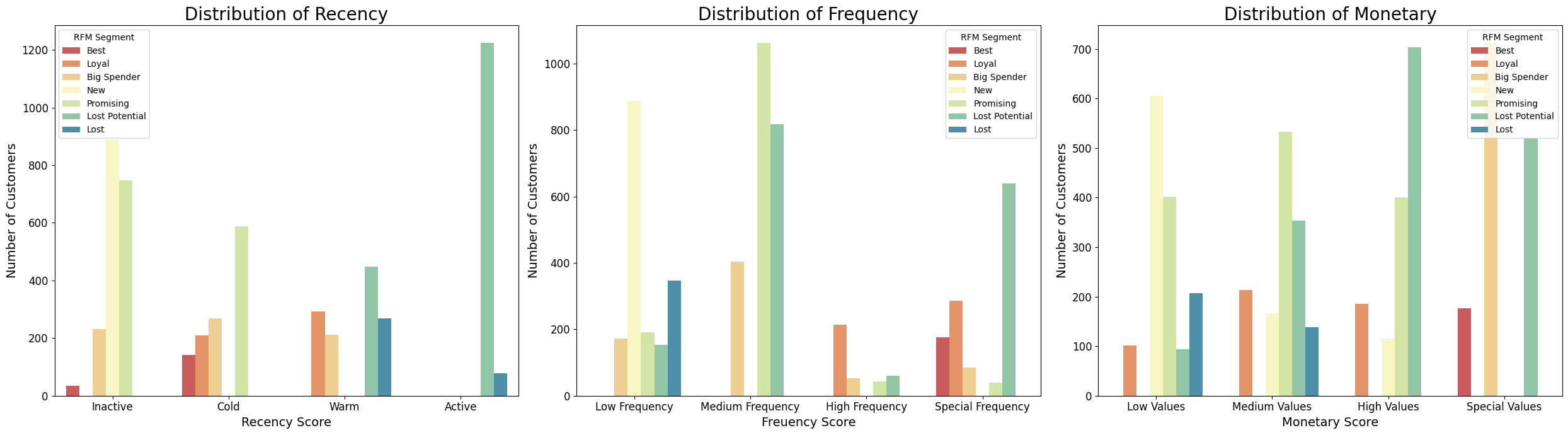}}
\caption{Distribution of Recency, Frequency, and Monetary scores by RFM segments.}
\label{fig:rfm-distribution}
\end{figure*}

The SHAP-based explanations are consistent with prior findings in e-commerce churn research \cite{ind12,ind13}. By understanding the key factors driving customer churn, e-commerce platforms can design personalized interventions to address specific issues and improve customer satisfaction, thereby reducing churn rates and increasing customer loyalty. This study indicates that churn risk decreases with greater tenure and higher cashback accrual, suggesting long-horizon levers such as loyalty and reward optimization. In the short term, complaint incidence shows the largest positive marginal effect. Thereby, prioritizing complaint handling should yield immediate results by lifting customer satisfaction

% \begin{center}
%     \includegraphics[width=0.6\linewidth]{shap-feature-importance.png}
%     \captionof{figure}{SHAP analysis result showing feature importance.}
%     \label{fig:shap-values}
% \end{center}

% \begin{center}
%     \includegraphics[width=0.6\linewidth]{SHAP - Value Distribution.png}
%     \captionof{figure}{SHAP analysis result showing value distribution.}
%     \label{fig:shap-values}
% \end{center}

\subsection{Survival Analysis for Customer Churn Prediction}\label{AA}
Survival analysis focuses on predicting the time of occurrence of a certain event, churn in our case. We model customer lifetime \(T\) and the survival function \(S(t)=\Pr(T>t)\), treating churn as the event and ongoing accounts as right-censored. The Kaplan–Meier estimator \cite{ind14} at ordered churn times \(t_{(1)}<\cdots<t_{(m)}\) is

\begin{equation}
\widehat{S}(t)=\prod_{t_{(j)}\le t}\left(1-\frac{d_j}{n_j}\right),
\end{equation}

where \(n_j\) is the number at risk just before \(t_{(j)}\) and \(d_j\) is the number of churn events at \(t_{(j)}\). The Kaplan-Meier survival curve (Fig. \ref{fig:km-survival-curve}) shows a steep early decline, from about 0.95 at onboarding to approximately 0.86\text{–}0.88 by month 6, followed by a slower decrease to roughly 0.76\text{–}0.78 by months 20\text{–}22, and a near-plateau through month 60. The median lifetime is not observed within the window since \(\widehat{S}(t)\) remains above 0.5. These dynamics indicate front-loaded churn in early tenure with greater stability thereafter. This indicates that customers who maintain engagement beyond this point tend to have a significantly lower risk of churning.

These findings align with previous literature where this temporal understanding of customer churn is used for effectively timing retention initiatives \cite{ind15,ind16}. Early months of a customer’s lifecycle represent a critical window to prevent churn, where efforts such as improved onboarding, encouraging repeat purchases, and early satisfaction interventions have the most impact. For longer-tenured customers, retention strategies typically shift towards rewarding loyalty and offering exclusive benefits to sustain engagement. 

\begin{center}
    \includegraphics[width=\linewidth]{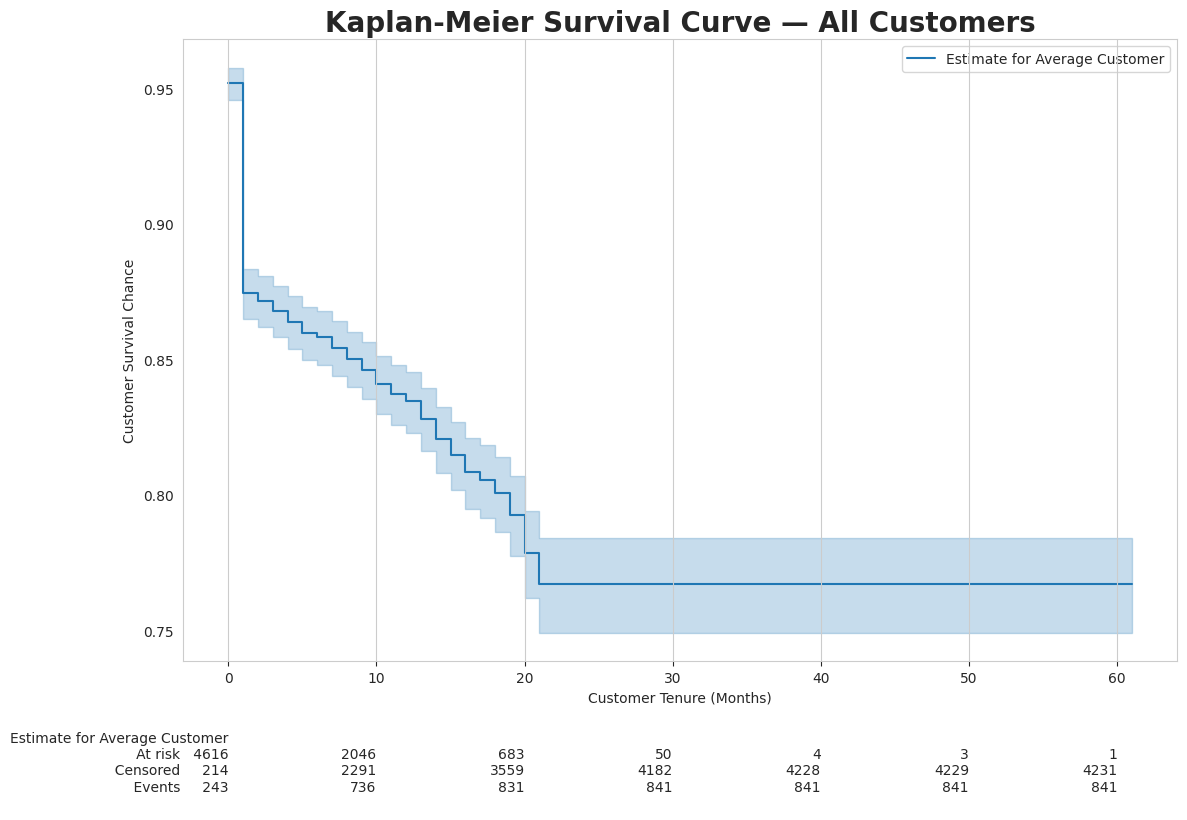}
    \captionof{figure}{Kaplan-Meier survival curve illustrating customer retention over time.}
    \label{fig:km-survival-curve}
\end{center}

\subsection{Customer Segmentation Based on Transactional Behavior}\label{AA}
Recency, Frequency, and Monetary (RFM) analysis is recognized for its ability to segment customers effectively and guide targeted marketing and retention strategies \cite{ind17,ind18}. Figure~\ref{fig:rfm-distribution} shows that \textit{Best} customers are recent, frequent, high-spend buyers, whereas \textit{Lost} customers are inactive with low frequency. Customers with high Recency, low Frequency, or low Monetary values face a greater risk of churn. Segmenting customers by RFM profiles supports personalized retention strategies.
\begin{itemize}
    \item \textit{New} and \textit{Promising} segments, with recent purchases but low frequency or spending, can benefit from personalized welcome campaigns, product education, or cross-selling offers to increase engagement.
    \item \textit{Lost} or \textit{Lost Potential} customers require personalized win-back initiatives such as discounts, targeted communication, or reactivation incentives to recover their interest.
    \item \textit{Best} and \textit{Loyal} customers may be engaged with loyalty rewards, early access to sales, or referral programs to reinforce their connection and maximize lifetime value.
\end{itemize}

% \begin{center}
%     \includegraphics[width=\linewidth]{RFM - Distribution.png}
%     \captionof{figure}{Distribution of Recency, Frequency, and Monetary scores by RFM segments.}
%     \label{fig:rfm-distribution}
% \end{center}

This threefold framework, combining explainable models, survival risk estimation, and behavioral segmentation, yields a comprehensive view of churn. TreeSHAP quantifies feature-level contributions, Kaplan–Meier curves reveal how risk evolves over tenure with censoring, and RFM-style segments summarize transactional patterns into actionable cohorts. Together, these outputs indicate why churn is likely, when intervention is most effective, and which customer groups to prioritize. This enables precise, time-aware retention strategies and more efficient allocation toward high-risk, high-value customers.

\section{Conclusion}
This study presented an integrated churn analytics framework that combines high-performing prediction with interpretability, temporal risk modeling, and behavior-based segmentation for e-commerce. Across five classifiers, XGBoost achieved the strongest performance and was therefore used for downstream analysis. TreeSHAP provided faithful local and global attributions, highlighting tenure, complaint history, distance to warehouse, and cashback as key drivers of churn propensity. Kaplan–Meier analysis characterized time-to-churn dynamics with an early high-risk window followed by stabilization, while RFM segmentation summarized transactional behavior into actionable cohorts. Together, these components explain why churn is likely, when interventions are most effective, and which customer groups to prioritize, enabling targeted, time-aware retention strategies.

\bibliographystyle{IEEEtran}
\bibliography{references}
\end{document}